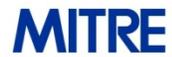

MTR200519
MITRE TECHNICAL REPORT

# TinyML for Ubiquitous Edge AI

**Bedford, MA**

**Author: Stanislava Soro**

**September 2020**

# Abstract

TinyML is a fast-growing multidisciplinary field at the intersection of machine learning, hardware, and software, that focuses on enabling deep learning algorithms on embedded (microcontroller powered) devices operating at extremely low power range (mW range and below). TinyML addresses the challenges in designing power-efficient, compact deep neural network models, supporting software framework and embedded hardware that will enable a wide range of customized, ubiquitous inference applications on battery-operated, resource constrained devices. In this report, we discuss the major challenges and technological enablers that direct this field's expansion. TinyML will open the door to the new types of edge services and applications that do not rely on cloud processing but thrive on distributed edge inference and autonomous reasoning.

This page intentionally left blank.

# Executive Summary

Tiny machine learning (TinyML) has emerged over the past few years as a new technological field at the cross-section of machine learning, embedded platforms, and software that leverages the power-constrained embedded platforms to enable embedded data analytics and enhance data processing on edge. Specifically, TinyML focuses on deploying machine learning and deep neural network models to the embedded, resource-constrained devices powered by microcontrollers (MCUs), thereby offering solutions for applications with low latency and low communication bandwidth constraints. The goal of this report is to understand the technology behind the TinyML framework, and to discuss the trends and challenges that TinyML technology currently faces.

In order for TinyML to become a mature technology, advancements across various fields need to take place. Research in chip design should include improving the existing chip solutions that will lead to more power-efficient and faster computations, and the design of novel chip architectures optimized to support specific neural network algorithms. Software libraries that provide computational blocks for efficient machine learning algorithm design have to be highly optimized for faster execution, yet agnostic toward the hardware platforms. Research in novel algorithm architectures aims to develop novel model compression methods to produce lightweight and reliable machine learning models that can run on embedded devices.

Furthermore, the challenges associated with cross-domain performance optimization need to be well understood. Fitting the neural network models to the resource-constrained hardware while considering the diversity across algorithms and hardware platforms is a particularly challenging problem. Within this problem, understanding power consumption vs. processing speed trade-off and its effect on the algorithm's accuracy is of particular interest for the TinyML community.

Beside challenges associated with the heterogeneity of hardware, software, and algorithms, there are challenges related to the process of model training, continuous model update, and remote deployment of these models to embedded devices. Currently, resource-constrained embedded devices are not capable of performing model training. The model training is done using a more powerful computing device, and then the pre-trained model is deployed to the embedded devices. In some cases, the challenges may arise when the model on the remote embedded device needs to be updated in the absence of a reliable connection to the device.

Several frameworks, which provide software to build TinyML models and speed up the application deployment, are currently in the development process. The frameworks discussed in this report (TensorFlow Lite [19], ELL [20], ARM-NN [21]) primarily focus on deploying neural network models to microcontrollers. They all integrate a similar workflow. A neural network model is first converted into a common format type and then optimized internally for deployment on the particular hardware platform. These frameworks are updated regularly by TinyML developers to support new machine learning algorithms and new hardware platforms.

TinyML will lead to novel applications across industry, military, and consumer spaces. Currently, the TinyML framework mostly focuses on applications involving computer vision, audio processing, or NLP algorithms. Applications in healthcare, autonomous systems, and surveillance will be supported by the TinyML framework in the near future and can be of particular interest to MITRE sponsors.

Despite the initial challenges, the expansion of TinyML technology will continue in various ways. Several companies specialized in chip design will continue the fundamental research on hardware optimization and application-specific chip design. The development of efficient software libraries for model optimization and seamless deployment will continue through engagement of the TinyML community in open source and proprietary TinyML frameworks. Finally, there are already many companies working on the development of efficient and pre-optimized solutions based on the TinyML software and off-the-shelf hardware, with the goal to shorten the final production cycle significantly. They all work together toward developing standardized machine learning models, practices, and benchmarking tools that will enable systematic development and broad adoption of TinyML technology.

# Table of Contents

# List of Figures

# List of Tables

This page intentionally left blank.

# 1  Introduction

Over the past few decades, data production has increased at unprecedented rates, caused by the wide adoption of sensing technologies and digitalization. This trend was also supported by the rise of data processing and storage capabilities provided by cloud technologies. The increase in computation resources accelerated the research and development of deep neural networks, which continued to grow in complexity and resource requirements over time. Today, cloud computing processes the largest AI models with millions of parameters that require gigabytes of memory and ultra-fast processing. These large neural network models focus primarily on accuracy and speed, as they have access to unlimited computational resources and memory.

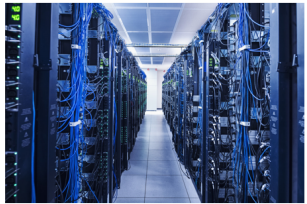  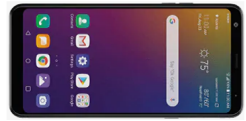  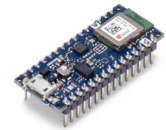

**Cloud ML (~2006)**
- DNN
- Large models (16-32GB)
- X Millions of parameters
- TFLOPs
- Focus on accuracy
- Hardware: GPU, TPU, FPGA…
- AlexNet, Inception, ResNet, VGGnet,…
- Data: storage, sharing (1%)

**Mobile ML (~2016)**
- Optimized algorithms, light CNN
- Constrained resources: memory 8GB RAM, application size limitation
- GFLOPs
- Focus on accuracy-efficiency trade-off
- Hardware: SoC, NPU
- MobileNet_v1, MobileNet_v2, ShuffleNet, SqueezeNet…
- Data: pics, audio, clicks, GPS (5%)

**Tiny ML (2019)**
- CNN-micro
- Severely constrained resources
- ~100KB RAM
- MCU with HW accelerators
- Sensors: CMOS cameras, IR cameras, audio, IMU, temp, chemical, accelerometers…
- Data: sensing the physical world (95%)

**Figure 1 Evolution of computing technologies.**

The emergence of mobile devices, such as laptops, smartphones, and tablets, gave rise to mobile computing. Mobile computing primarily refers to the field of wireless communications for carry-on mobile devices, which benefited from technical advances such as low-power processors, compact size memory, and low-power display technology. Soon, mobile computing platforms started to be used in many fields such as robotics, self-driving cars, or augmented reality, which requires the computation of real-time tasks in a very short time, without relying on the computing power from a cloud. This shift toward mobile platforms initiated the development of a new generation of neural networks [1], [2] that are more compact in size and focus on model efficiency, besides model accuracy.

In parallel to the expansion of machine learning and AI fields, the IoT field expanded too, propelled by the advancements in connectivity and abundance of system-on-ship solutions. Today, the number of microcontroller (MCU)-powered devices is growing exponentially. There are around 250 billion microcontrollers deployed today and is estimated that about 40 billion microcontrollers will be deployed in the next two years [3].

However, in many cases, embedded devices do not process any collected data; instead, the data is transmitted to a remote location for storage and further processing. In some application, this can cause undesired data latency, data loss, and data privacy issues. This prompted research toward understanding the fundamental challenges in deploying machine learning algorithms to embedded devices with limited resources (in terms of processing speed, memory, energy).

Embedded devices usually collect data from various sensors (CMOS cameras, IR cameras, audio, IMU, temperature sensors, chemical sensors, accelerometers, etc.). These embedded devices are often battery-powered and run in a low power range (1 mW or less). In many cases the microprocessor on the device stays underutilized over time, since it usually performs simple processing tasks. The widespread use of embedded MCU-powered devices, their proximity to the physical sources of information through sensors, and underutilized processing capabilities make them highly suitable for the deploying lightweight inference algorithms. This trend gave rise to tiny machine learning (TinyML), which is an emerging embedded technology field that aims to deploy machine learning algorithms to resource constrained, MCU powered devices.

The TinyML community addresses the fundamental challenges of enabling sophisticated machine learning algorithms to operate under heavily constrained resources of embedded devices and performing highly accurate data understanding tasks. To picture this challenge, we refer to Figure 2 that shows the top-1 accuracy of various deep neural network models developed over the past decade. It is important to understand that these models were developed using significant processing resources available at time. The Figure 2 shows that the model accuracy increases as a function of model complexity (presented by the size of a "ball") and computational complexity. The goal of TinyML is to find ways to adapt these deep learning algorithms for use on MCU-based embedded platforms with significantly fewer resources and to develop supporting practices that will enable easy deployment and high accuracy of deployed models.

TinyML will enable innovations in various fields, such as distributed cyber-physical systems, autonomous systems, healthcare, consumer electronics, and in the general field of artificial intelligence. However, achieving these goals calls for joint solutions from many disciplines, including but not limited to machine learning, computer architecture, signal processing, optimization, and hardware design. The development of TinyML requires collaborative effort between embedded systems community and machine learning community to bring together people with different areas of expertise to establish the directions and work on challenges of this highly experimental, multi-disciplinary field. In this report, we evaluate the current state of the art of the TinyML framework that aims to bringing AI to the smallest, least powerful edge devices.

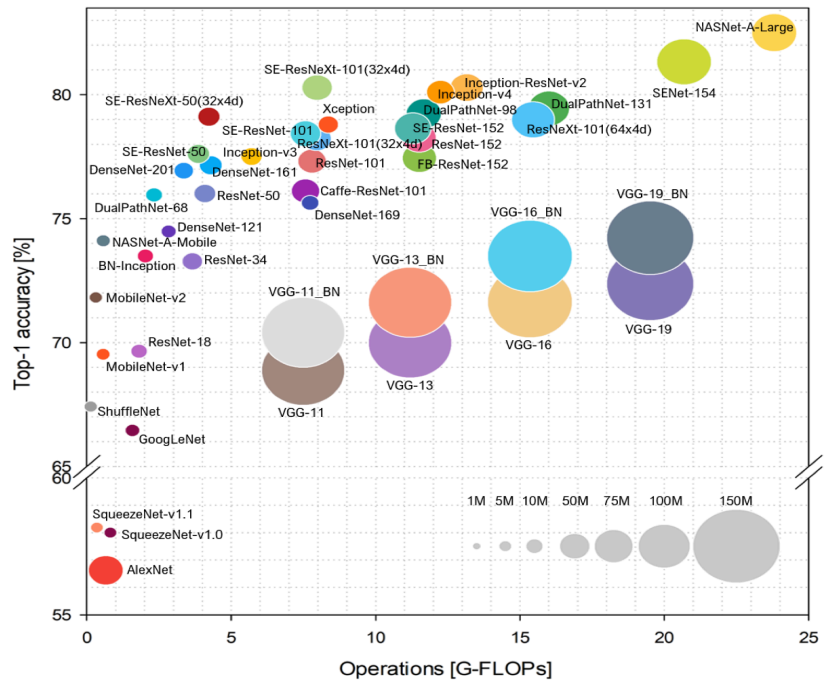

**Figure 2 Top-1 accuracy vs computational complexity for various deep neural network models developed over the past decade [4].**

## 2 Machine Learning on MCUs

There are several advantages of performing data analytics on the embedded MCU-powered edge device, including:

*Latency* – Latency is a crucial requirement for time-critical applications. Data transmission to the cloud typically introduces longer latency than data processing on the embedded device. Applications built around user experience and automotive safety applications are examples where latency cannot be tolerated, so data must be processed on the device.

*Reduced data transmissions* – When the network connectivity is unreliable or even nonexistent, data processing must take place at the edge device. It is often impractical to stream data from sensors that generate large amounts of raw data, such as cameras, accelerometers, or microphones.

In event-based applications, data analytics on the edge device can filter out invaluable data before sending out the processed information.

*Energy efficiency* – MCUs are inherently designed to perform extremely low-power computations, so they can operate for a long time on a small power budget. MCUs can run on batteries or other types of energy sources, such as energy harvesting.

*Low cost* – The expansion of IoT applications in many industrial segments was driven by the low cost and high volumes of microprocessors and supporting embedded hardware, and this trend is likely to continue.

*Privacy* – Privacy is always a concern when personal or other highly sensitive data is handled outside of the device. This is particularly true if data is sent to the cloud, as data can be lost or stolen. Thus, data processing on the device results in better data privacy and better data protection.

*Autonomy* – Small embedded devices can contribute to the increased autonomy of the system they belong to, by generating independent, data-driven local decisions that can be shared only within the system.

However, running machine learning on embedded 32-bit processors entails many technical challenges. We will discuss these challenges next.

## 2.1 Challenges of Machine Learning on Embedded Devices

Running complex machine learning models, such as neural networks on highly constrained embedded hardware requires careful software and hardware co-design. The two key challenges in deploying neural networks on microcontrollers are the small memory size and the short battery life [5].

The size of the machine learning model has to be small enough to fit within the constraints of the MCU-powered device. As illustrated in Table 1, typical microcontrollers have extremely limited on-chip (SRAM) memory (192–512 KB) and flash storage (256 KB–2 MB). The entire neural network model with its weights, neural connections, and supporting code has to fit within the small flash memory. The SRAM size limits the temporary memory buffer used to store the model's input and output activations. These constraints need to be addressed during the design of compressed models that perform in real time with high accuracy.

| MCU Platform | Processor | Frequency | SRAM | Flash |
|---|---|---|---|---|
| Arduino Nano 33 BLE Sense [6] | ARM Cortex M4 | 64 MHz | 256 KB | 1 MB |
| ESP32 [7] | Tensilica Xtensa LX6 | 160 MHz | 512 KB | 2 MB |
| Sparkfun Edge Appolo3 Blue [8] | ARM Cortex M4F | 48 MHz | 384 KB | 1 MB |
| ST Nucleo Boards [9] | ARM Cortex M7 | 216 MHZ | 320 KB | 1 MB |
| Adafruit EdgeBadge [10] | ATSAMD51 | 120 MHz | 192 KB | 512 KB |

**Table 1 Representative devices supported by TensorFlow Lite for Microcontrollers.**

The typical application use cases commonly assume the sporadic events (such as when the object of interest is present/not present), which further constrains the system design by requiring that the peak memory usage during the detected event does not exceed the limited SRAM memory. Peak memory usage is the maximum amount of total memory used during inference, including memory needed to store the neural network's activation maps. To perform fast inferences, neural network activations should be preferably stored in the fast SRAM memory. If large transient activations require space beyond SRAM they can be stored in slower off-chip memory, typically causing long wait times [5].

To estimate peak memory usage, the model must be analyzed in terms of multiply-add operations per inference. Then, considering the speed of the microprocessor (typically in the range of 100–200 MHz) and the required number of inferences per second, one can calculate the limit for multiply-add operations per inference. This process is used to estimate how fast the deployed neural network model runs on a particular embedded device and how well it can support the targeted application.

Another limitation of running TinyML on the microprocessor is limited battery life. To estimate battery life, the energy efficiency of the microcontroller should be considered. In general, energy efficiency depends on the machine learning algorithm's computational cost and the processing duty cycle. There are many challenges with measuring energy consumption, as different devices can consume drastically different power, making it hard to compare them. Also, often it is hard to understand what falls under the power measurement scope, as different devices use different ways of data preprocessing or additional peripherals that impact energy consumption during the measurements.

# 3 Overcoming Challenges of Machine Learning on the Edge

There are several approaches to overcome the challenges of limited memory and low-speed processing on embedded devices. Some of them include:

*Model reduction* – This is an active research area where the goal is to reduce the size of the neural model to fit it to the microcontroller. Several methods are used, such as model shrinking (reducing the number of network layers), model pruning (setting the low-value weights equal to zero), or parameter quantization.

*Lightweight frameworks* – Several frameworks are developed by industrial companies and research institutions that provide open source tools for designing and implementing machine learning algorithms on resource-constrained devices. These frameworks include highly efficient inference libraries and workflow processes that simplify the model development and model deployment process to the embedded device.

We will further discuss the techniques for model reduction and software frameworks in more details below.

## 3.1 Model Compression Algorithms

Neural networks were able to achieve dramatic accuracy improvements over the past couple of years, thanks to large models with millions of parameters and the availability of high computing resources. However, these models cannot be directly deployed to the embedded devices with low memory. Over the past few years, significant progress has been achieved in model compression techniques. Model compression combines methods from various fields such as machine learning, signal processing, computer architecture, and optimization, aiming to reduce the model size without significantly reducing model accuracy.

Model compression includes several techniques that can be represented as an operational pipeline process, as illustrated in Figure 3, where operations are applied successively on the neural network model, resulting in reduced model size [11]. We discuss these operations in more detail next.

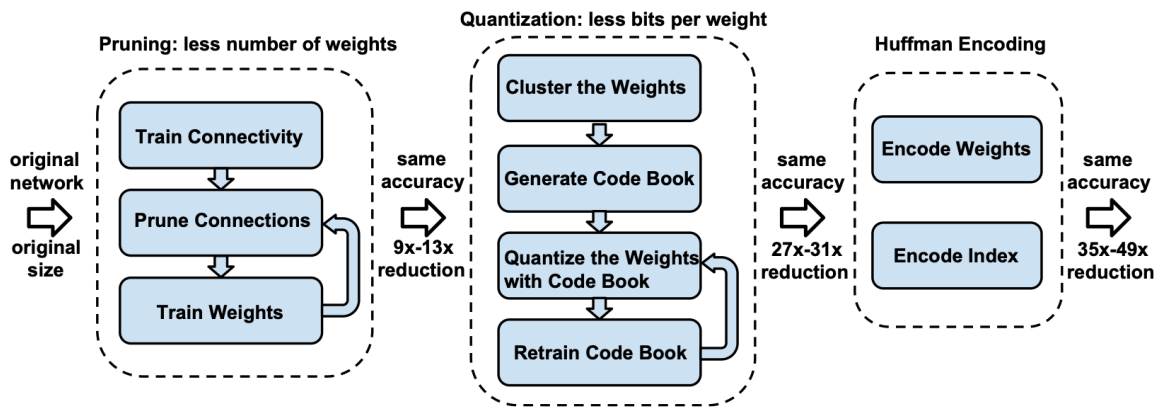

**Figure 3 The pipeline processing architecture introduced in [12] for model reduction. Each block introduces a significant reduction in model size with no effect on model accuracy.**

### 3.1.1 Model Pruning

The model reduction process starts with model pruning, aiming to reduce the number of connections in the neural network model by pruning non-informative weights based on some loss function. Recent work presented in [12] describes an example of the pruning process. Model pruning starts with the regular network training, where the connections between the layers are learned, followed by the pruning of small weight connections (usually using fixed threshold value). After this, the network can be retrained to learn the final weights of the remaining sparse connections again. Many other pruning methods are proposed in the literature, such as weighs pruning based on Hessian of the loss function [13] or using the percentage of zero outputs to prune unimportant connections [14].

### 3.1.2 Parameter Quantization

The next step involves the network quantization process, where the network size is reduced further by reducing the number of bits needed to represent each weight. The quantization process involves deciding the number of bits needed to represent weights based on observed [min, max] range of the weight values. This is followed by deciding the best ways to allocate bits for the integer part of the weights to cover the whole quantization range, considering that some min or max values may be cut off. For example, one way of doing it is first to calculate mean ($\mu$) and standard deviation ($\sigma$) of all weights, and then select the number of bits to allocate for the representation of the integer part so that it is possible to represent all values in the range [$\mu+3\sigma$, $\mu-3\sigma$] [15]. There are many other methods in the literature proposed to quantize the weights efficiently, ranging from linear quantization, non-linear (log based) quantization, and k-means clustering, to even single bit quantization for binary-weight neural networks.

### 3.1.3 Network Compression

Compression of network coefficients using lossless coding is discussed in [12], where Huffman coding is applied to quantized weights for further compression, which resulted in the reduction of 20%–30% in memory space needed for network parameter storage.

### 3.1.4 Knowledge Distillation

Knowledge distillation is another way of extracting knowledge representation into a smaller form. Proposed in [16], it refers to the set of methods used to transfer knowledge from a "teacher" model to the much smaller "student" model. It is based on the fact that a smaller model does not have a sufficient capacity to learn all interdependencies of a large dataset so that this knowledge can be transferred from the larger model. The student model is represented as a shallow version of its parent model, and it is trained based on softmax[1] output from the parent model using a penalized function. Although the knowledge distillation method produces compressed neural network models, it is limited to models with softmax output. Also, this approach still underperforms compared to other model compression methods. Knowledge distillation techniques are useful when the application involves small data sets or when it requires a significant efficiency improvement.

## 3.2 Model Selection: Complexity-Accuracy Trade-off

The neural network model design starts with careful selection of the initial network model architecture according to the application, which is then subsequently refined during the design process to achieve required model accuracy within limited resource constraints of the embedded device [17].

For example, Table 2 shows various models used for different audio tasks, ranging from the more straightforward task, such as voice activity detection, to the more complex ones, such as speech recognition. According to the table, simple tasks can be accomplished using fully connected layers (FC) and fewer parameters. In contrast, complex tasks are accomplished using recurrent neural networks with LSTM units and many parameters. Often, the conditions under which the model will operate after deployment need to be taken into account during the network design process. For example, a model for an audio task needs to be robust for different SNR conditions under which a network model may operate, affecting the size of the resulting model.

---

[1] Softmax function is often used as the last activation function of a neural network to normalize the output of a network to a probability distribution over predicted output classes.

| Task | Network Type | Network Architecture | Number of Parameters |
|---|---|---|---|
| Voice activity detection [23] | MLP[2] | 60-24-11-2-FC[3] | 5 K |
| Keyword spotting [26] | CNN[4] | 1CL[5]-FCL-3-FCNL[6] | 54 K |
| Speaker recognition [27] | CNN | 1CL-3-FCNL | 234 K |
| Speaker verification [24] | RNN[7] | 2x220 GRU[8] | 900 K |
| Speech enhancement [24] | RNN | 500-1024-1024 FC | 500 K |
| Speech recognition [25] | RNN | 5x465 GRU | 10 M |

**Table 2 Various audio tasks and corresponding implementations using various neural network models [17]. Task complexity reflects the size of the neural network model and the number of model parameters.**

Understanding of differences and trade-offs between different neural network architectures is critical for the model selection process. For example, multi-layer perception networks usually have a small footprint and are easy to train; however, they often exhibit limited performances in terms of model accuracy. Convolutional neural networks are generally easier to optimize for good performance, and they are successful in low-level feature extractions. On the other side, they often require a large number of operations to converge to a good result. They also use fixed input size data blocks, which is not the best option for processing data that involves temporal dependency, such as audio processing.

Recurrent neural networks consider time and therefore perform well on tasks that involve temporal sequences, which makes them particularly suitable for audio tasks. However, the recurrent models usually contain large number of parameters, making them more challenging for deployment on highly constrained embedded devices.

## 4  TinyML Frameworks

Due to limited resource constraints, MCU-powered devices cannot currently support model training. Instead, the model is first trained in the cloud or on a more powerful device, and then deployed back to the embedded device. However, since it is expected that the number of deployed smart devices will increase exponentially, so will the on-device training become critical for ensuring timely updates of application and security software on embedded devices. A good overview of challenges and directions toward making on-device training possible is provided in [34].

There are three ways to deploy the model to the embedded system: coding by hand, code generation, and ML interpreters [18]. Coding by hand allows for low-level application-specific optimizations; however, it is time-consuming and impractical from the standpoint of limited ability to share knowledge and adopt new shared methods, which may negatively affect the adoption pace of TinyML. Code generation produces the well-optimized code without the significant hand coding effort, by abstracting and automating system-level optimizations. However, each vendor uses its proprietary tools and compilers, which makes portability and comparability a challenge. Finally, ML interpreters are made for portability, as they provide the same abstract structure across different platforms. An ML interpreter is usually a part of a framework that encompasses a set of tools and software libraries for implementing machine learning algorithms on embedded devices with limited processing capabilities, including MCUs. Within a framework, an interpreter is used to call individual kernels (such as convolution, padding, etc.), which can be individually optimized for the particular platform.

---

[2] MLP – Multi-layer perception network

[3] FC – Fully connected layer

[4] CNN – Convolutional neural network

[5] CL – Connected layer

[6] FCNL – Fully connected layer with nonlinear function

[7] RNN – Recurrent neural network

[8] GRU – Gated recurrent unit

The term "TinyML" is often used as a synonym for TensorFlow Lite [19], a framework developed by Google that focuses on machine learning implementation on mobile embedded devices. However, other companies and research institutions have developed several other frameworks, with the intention to promote their hardware platforms or software libraries. Some of the most popular TinyML frameworks are discussed next.

It is interesting to note that these frameworks focus on neural network models only. Several other open-source tools include libraries for other machine learning algorithms, such as naïve Bayes, decision trees, and k-NN. A comprehensive list of these tools is given in [22].

## 4.1 TensorFlow Lite

TensorFlow Lite is an open-source deep learning framework developed by Google for inference on embedded devices [19]. This framework consists of two main tools, Converter and Interpreter. TensorFlow Converter transforms the TensorFlow code into a particular format, reduces the model's size, and optimizes the code for minimal accuracy loss. TensorFlow Interpreter is a library that executes the code on the embedded device.

A separate port of TensorFlow Lite, called TensorFlow Lite for Microcontrollers, is designed to run machine learning models specifically on 32-bit microcontrollers with just a few kB of memory. It was successfully ported to many processors, including the well-known Arm Cortex-M Series and ESP32. Figure 4 illustrates the steps needed to build "wake words" application using this framework. The framework is available as an Arduino library so that it can be ported to other C++ 11 projects as an open-source library. The MCU's constraints cause some of the limitations of TensorFlow for Microcontrollers. For example, the framework supports only a subset of TensorFlow operations, and it cannot be used for model training. The full list of currently supported devices is given in [19].

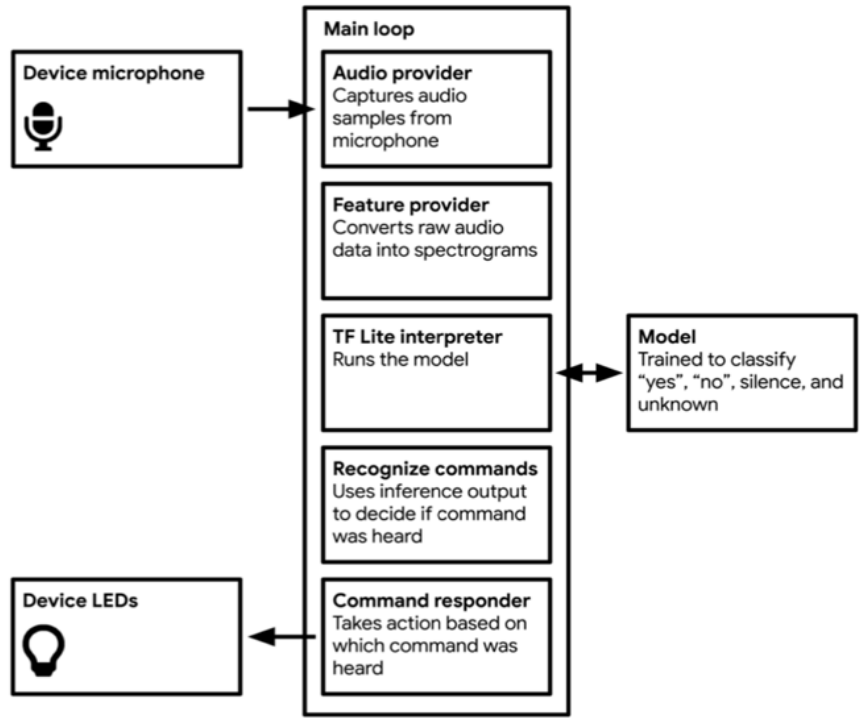

**Figure 4** An example of wake words application developed using TensorFlow Micro [19] tool.

## 4.2 Embedded Learning Library (ELL)

Embedded Learning Library is an open-source embedded framework developed by Microsoft that supports the development of machine learning models for several platforms based on ARM Cortex-A and Cortex-M architectures, such as Arduino, Raspberry Pi, and micro:bit [20].

ELL can be described as a cross-compiler toolchain for ML/AI models. It compresses the model and generates optimized machine-executable code for an embedded target platform. The input into the ELL system is a model generated in various formats, such as OpenNeural Network Exchange (ONNX) format [30], TensorFlow [31], or Microsoft Cognitive Toolkit (CNTK) [32] format. The ELL system generates an intermediate model as a .ell file. The intermediate .ell model file is used to generate executable code. ELL is an ambitious project that aims to be agnostic with respect to the format of the input model. However, at the time of writing this report, it is supported by a smaller team, and it is not clear how widespread this toolkit is.

## 4.3 ARM-NN

The ARM-NN [21] is an open source Linux software for machine learning inference on embedded devices, developed by Arm. The core of Arm's framework is the CMSIS NN [33] software library, which is a collection of efficient neural network kernels developed to maximize the performance and minimize the memory footprint of neural networks on Cortex-M processor cores. Since CMSIS-NN specifically targets embedded devices, it uses fixed-point arithmetic, so the model parameters are quantized to 8-bit or 16-bit integers and then deployed to the microcontroller for inferencing. Also, this framework exploits processing architectures of Arm's most popular Cortex-M series of microcontrollers, in order to improve the execution time.

# 5 TinyML Applications

Enabling machine learning/AI algorithms on embedded devices will lead to innovation across industry, consumer, and military spaces. As TinyML research and development accelerate across industry and academia, the list of domains and new applications will expand fast. Adding a layer of intelligence to embedded devices will proliferate the applications in domains that significantly depend on IoT and embedded technology, including healthcare, industrial monitoring, and consumer electronics. This section discusses some of the domains where TinyML applications are expected to grow in the near future.

## 5.1 Healthcare

TinyML has great potential to enhance various monitoring and personal health products. Wearable devices require a significant power budget for continuous sensing and streaming of a person's physiological and activity data, which requires uninterrupted connectivity and privacy protection. Compact, pre-trained inference models on device for signal denoising, temporal analysis, and classification based on TinyML can provide a comprehensive assessment of collected personal data in real-time, effectively avoiding the continuous data streaming. The development of low-power, highly accurate, real-time inference algorithms for wearable devices will be especially crucial for more complex, data-rich physiological sensors, such as ECG, which still pose a significant challenge for today's wearable technology.

Several companies explore TinyML-type frameworks to enhance various personal health products, such as hearing aids [29]. Similar trends are expected for other personal health and well-being applications, such as vision enhancement or gait tracking. Multi-sensor data fusion and deep neural network inference on an embedded device will enable continuous monitoring and assessment of a person's well-being and mental health, thereby enhancing the ways to treat people with various mental conditions, such as dementia, depression, or post-traumatic stress disorder.

## 5.2 Surveillance

Currently, TinyML focuses mostly on developing neural network algorithms for computer vision and audio processing, which are the foundation of surveillance applications. Object detection, person detection, face detection, simple activity recognition, and voice activity detection are examples of tasks that can be performed with high accuracy by running TinyML-based code on MCU-powered devices. TinyML will enable the development of new types of ubiquitous, lightweight, and low-cost surveillance, monitoring, and identification applications. It is particularly suitable for applications where a simple, classified answer is needed (such as person detected or not) as opposed to providing full contextual information.

## 5.3 Embedded Security

Cyber-physical systems and IoT are increasingly becoming the targets of cyber-attacks. Embedded security is becoming one of the fastest-growing fields that addresses the vulnerabilities of these systems. Data analytics on edge devices is becoming crucial in distinguishing between normal and unexpected network traffic patterns, understanding the type and souce of threat, and addressing cyber-attacks. The embedded device may run pre-trained machine learning models to detect the anomalies in network traffic patterns (such as unexpected activity levels over time, suspicious traffic types, unknown origin/destination, etc.). Previously undiscovered attacks, for which the model does not have an existing pattern, would be discovered this way. Similarly, software-level vulnerabilities may be addressed by building a learning model to recognize code changes or file changes, which would indicate the need for additional code patching. Distributed systems may exploit novel machine learning approaches, such as federated learning [35] supported by TinyML framework, to build a general model of threat of a cyber-physical system. Detection of device tampering and unauthorized physical access can be detected using embedded vision or audio processing based on TinyML.

## 5.4 Industrial Monitoring

We are witnessing the rapid development of Industrial Internet practices, seen through advancements in the automation of traditional industrial segments, enhanced connectivity, and smart sensing technology. TinyML technology is well-positioned to support further enhancements of these practices. In many industrial monitoring applications, predictive data analytics can be implemented on embedded devices. TinyML can address the needs for in-place data analytics and time-critical constraints of some industrial monitoring applications. Indoor localization services for asset tracking can be designed based on using neural network models with RF signal inputs that can be periodically retrained using collected RF data for better location accuracy.

## 5.5 Autonomous Systems

Various autonomous systems, such as robotics or autonomous cars, use edge computing to improve performance and safety. TinyML that brings edge computing tasks to small, lightweight, power constrained MCU devices can address design issues of these systems related to restricted power budget or small form factor.

## 5.6 Augmented/Virtual Reality

AR/VR technologies are growing at a rapid rate. These technologies are built around the idea of getting closer to the user by offering enhanced user experience, through various applications from personal assistance, environment sensing, and gaming, to social interaction. The AR/VR technology has to support a much wider range of input and output modalities while considering constraints such as limited power budget, small form factor, and limited bandwidth for data streaming. Leading AR/VR companies are currently looking into using embedded computer vision and TinyML framework to address some of these issues.

## 5.7 Smart Spaces

The goal of smart space applications is to provide services to a user through their interaction with technology that senses the user's physical environment. The state of the environment and the user are perceived using various sensing technologies and actuators. Smart services should have a high degree of autonomy and adapt to the user's immediate needs with minimal user intervention. TinyML technology can take part in many tasks of smart space applications while preserving the user's privacy. For example, embedded computer vision and audio processing can be used for user identification or user activity recognition tasks, such as gesture recognition that can initiate the actions in a smart space environment (e.g., door opening/closing, turning the lights on/off, temperature adjustment).

# 6 TinyML Challenges

TinyML is in a very early stage, and the research and development in this field is developing rapidly. Research in this field spans many directions, and its success will depend on the breakthroughs at the intersections of various fields, including machine learning algorithms, computer architecture, and hardware design. Along this way, there are many challenges that this technology faces, including:

*Hardware and software heterogeneity* – Similar to IoT space, there is vast diversity in the type of hardware components and algorithms in the TinyML space, making it hard to understand the trade-offs between different TinyML implementations.

*Lack of benchmarking tools* – Embedded systems running TinyML-based algorithms need to be compared systematically in order to gain a deep understanding of the TinyML performance. A set of benchmarking tools and methods is required to evaluate, compare, and systematically capture various performance differences between systems. The TinyML community has recently established an organization called TinyMLPerf [28] to provide guidelines and procedures for benchmarking of TinyML systems considering various aspects, such as performance, power consumption, memory usage, and differences in hardware. An excellent overview of challenges associated with comparing different hardware platforms used for TinyML applications is provided in [18].

*Adaptation and lifelong learning* – In most cases, the software on MCU-powered low-power devices is rarely updated. Once programmed, these devices run the same software during their lifetime. However, as these devices collect data about new events that are not a part of the initial training dataset, the deployed inference models should evolve and reflect new conditions. The need for continuous model update poses a question of integrating new learnings into the evolving model and how to perform the actual model update, as updating the software on many remote devices may be a challenging, and sometimes impossible task.

*Lack of appropriate datasets* – Although there are many well-known public datasets available for training machine learning models, there is still a lack of generally adopted datasets to train ultra-low-power models for embedded devices. Ideally, these datasets should reflect temporal and spatial resolution, noise level and diversity of data that corresponds to data expected from sensors attached to low-power embedded device. Having a standard set of datasets is crucial for systematic development of general practices and algorithms for embedded inference and system performance benchmarking.

*Lack of widely accepted models* – The widespread adoption of machine learning for the MCU-based class of embedded devices will be possible once the standard, generally accepted set of models is widely available. Similar to MobileNet [1], which became a baseline model for benchmarking different neural network models for mobile edge computing devices, a set of generally applicable TinyML-based learning models developed by the TinyML community can accelerate adoption of TinyML framework.

*Need for other types of machine learning models* – Currently, the TinyML community is mainly focusing on the development of deep neural network models, thanks to their superior performance. However, the framework should explore other types of machine learning algorithms (such as ensemble models, SVM, decision trees, etc.) due to their lower complexity and smaller resource requirements.

# 7  Conclusions

Tiny machine learning (TinyML) has the potential to unlock an entirely new class of smart applications across industrial and consumer spaces. This field is still at a very early stage of development, but it evolves rapidly. Currently, the focus of TinyML is on understanding the performance boundaries and trade-offs among different integral parts (machine learning algorithms, hardware, software) of resource-constrained embedded systems.

The edge computing landscape is broad, and TinyML deployments are often highly customized. The diversity of hardware and software needs to be addressed through benchmarking policies for TinyML to gain wide support.

TinyML will enable novel research directions, focusing on understanding how inference at the edge can impact other aspects of the complex systems, such as connectivity, autonomy, and system resilience to cyber-attacks. It has great potential to enable research and development of the machine learning field from a unique perspective, which can unleash innovative solutions in many domains.